\documentclass[preprint]{elsarticle}

\usepackage{lineno,hyperref}
\usepackage{breakurl}
\usepackage{booktabs}       

\begin{document}

\begin{frontmatter}

\title{Feature representations useful for predicting image memorability}

\author{Takumi Harada\corref{mycorrespondingauthor}}
\ead{t-harada@mosk.tytlabs.co.jp}
\author{Hiroyuki Sakai}
\address{TOYOTA CENTRAL R\&D LABS., INC., 41-3 Yokomichi, Nagakute, Aichi, Japan}

\cortext[mycorrespondingauthor]{Correspondence}

\begin{abstract}
Prediction of image memorability has attracted interest in various fields. Consequently, the prediction accuracy of convolutional neural network (CNN) models has been approaching the empirical upper bound estimated based on human consistency. However, identifying which feature representations embedded in CNN models are responsible for the high memorability prediction accuracy remains an open question. To tackle this problem, we sought to identify memorability-related feature representations in CNN models using brain similarity. Specifically, memorability prediction accuracy and brain similarity were examined across 16,860 layers in 64 CNN models pretrained for object recognition. A clear tendency was observed in this comprehensive analysis that layers with high memorability prediction accuracy had higher brain similarity with the inferior temporal (IT) cortex, which is the highest stage in the ventral visual pathway. Furthermore, fine-tuning of the 64 CNN models for memorability prediction revealed that brain similarity with the IT cortex at the penultimate layer positively correlated with the memorability prediction accuracy of the models. This analysis also showed that the best fine-tuned model provided accuracy comparable to state-of-the-art CNN models developed for memorability prediction. Overall, the results of this study indicated that the CNN models' great success in predicting memorability relies on feature representation acquisition, similar to the IT cortex. This study advances our understanding of feature representations and their use in predicting image memorability.
\end{abstract}


\end{frontmatter}

\section{Introduction}
The prediction of image memorability has attracted interest in various research fields including computer vision and neuroscience. Defined as the extent to which an image is considered memorable, image memorability is measured using a visual memory task \cite{ICCV15_Khosla,6629991}. Its prediction has the potential for a wide variety of applications, for example, creating textbook illustrations that stick in the learner's mind based on image memorability \cite{oliva2013makes}.

Recent studies in computer vision have used convolutional neural network (CNN) models, and a high prediction accuracy for image memorability has been achieved. Perera et al. \cite{Perera_2019_CVPR_Workshops} examined whether feature representations in a CNN model for object recognition were useful for predicting memorability. They found that the prediction accuracy was comparable to the empirical upper bound estimated based on human consistency (inter-individual variability in visual memory performance). Similar results have been reported using different CNN models \cite{fajtl2018amnet, praveen2021resmem}. These findings clearly indicate that image memorability can be successfully predicted using CNN models.

However, it remains unclear which feature representations embedded in the CNN models are responsible for the high prediction accuracy of image memorability. Although some attempts have been made to examine the relationship between memorability and low-level image features, such as color, no clear explanations have been found so far \cite{6629991,7410487}. A comprehensive CNN model analysis is required to identify feature representations useful for memorability prediction. Understanding such feature representations can provide insights into the design of network architecture \cite{NEURIPS2019_70117ee3, NEURIPS2020_98b17f06}.

A plausible approach for exploring feature representations in CNN models is investigating brain similarity. The usefulness of this approach was proven by Yamins et al.'s pioneering study \cite{yamins2014performance} that showed representational similarities between the visual cortex and CNN models for object recognition. This finding suggests that brain-like feature representations in CNN models benefit object recognition. Recently, Schrimpf et al. \cite{schrimpf2018brain} proposed a benchmark called Brain-Score, capable of assessing the similarity of internal feature representations between CNN models and visual cortices in the ventral visual stream. Using this benchmark, they replicated the higher model performance for object recognition and higher representational similarity with the inferior temporal (IT) cortex, the last stage in the ventral visual stream. This result is consistent with biological evidence that the IT cortex plays a crucial role in object recognition \cite{dicarlo2012does}. This brain similarity approach might be more suitable for identifying feature representations for memorability because image memorability, unlike object categories, inherently relies on the brain's neural processing. However, no such attempts have been made to date.

This study investigated memorability-related feature representations in CNN models in terms of brain similarity. We hypothesized that the high prediction accuracy of image memorability with CNN models is underpinned by feature representations similar to the IT cortex, based on recent neuroscience evidence that population response magnitude in the IT cortex was correlated with monkey memorability performance \cite{jaegle2019population}. To validate this, we examined the relationships between memorability prediction accuracy and brain similarity across 16,860 layers in 64 object recognition CNN models. In addition, we assessed the feature representations of the penultimate layers in the 64 CNN models fine-tuned to an image memorability dataset.

%
%
\section{Materials and Methods}

\subsection{CNN models}
A total of 64 CNN models were included in the current analysis (Table S1). The ImageNet dataset was used to train all models that were previously developed for object recognition \cite{deng2009imagenet}. They contain various network structure types, such as shallow (e.g., AlexNet \cite{krizhevsky2012imagenet}, VGG16 \cite{simonyan2014very}, and ResNet18 \cite{he2016deep}) and deep (e.g., DenseNet201 \cite{huang2017densely} and SE-ResNet152 \cite{Hu_2018_CVPR}) networks. All models were proposed to achieve higher object recognition accuracy, except for CORnet \cite{kubilius2018cornet}, which was developed for exploring the neural mechanisms of object recognition. All models are available in \cite{pytorch, cadena} under a BSD 3-Clause License, except for CORnet \cite{cornet} under a GPL-3.0 License and EfficientNet \cite{eff} under an Apache-2.0 License. 

\subsection{Datasets}
The large-scale LaMem \cite{ICCV15_Khosla} image dataset was used for memorability, which contains approximately 60,000 images, each labeled with a memorability score determined in a psychological experiment in which participants were asked whether a given image had been seen before during the experiment. The hit rate at which the image repeat was correctly detected was obtained for each image and averaged across participants. The average hit rate with a modification that considers time delay was defined as the memorability score for the images (see \cite{ICCV15_Khosla} for more details). The score distribution in the dataset was between 0.2 and 1.0. It is known from previous reports \cite{ICCV15_Khosla,6629991,goetschalckx2019memcat} that the average Spearman's correlation coefficient $\rho$ among subjects is in the 0.68--0.78 range. The average Spearman's correlation coefficient between the subjects in the LaMem dataset was 0.68. The images it contains were sampled from various datasets and included a variety of objects comprising landscapes, objects, art, emotionally evocative images, and selfies.

\subsection{Brain similarity}
Brain-Score is used to evaluate the representational similarity of CNN layers with visual cortices (called here “brain similarity") \cite{schrimpf2018brain}. This similarity index was calculated as the prediction accuracy (Pearson's correlation coefficient) of neural responses to various images in each visual cortex (V1, V2, V4, or IT) with the CNN layer's outputs in response to the same images (depicted using orange dashed lines in Figure 1). The prediction was made using partial least squares (PLS) regression with the number of latent variables set at 25 and employing 10-fold cross-validation for generalization. The brain similarities of 16,860 layers in the 64 CNN models were computed after excluding layers with fewer than 24 units. This analysis was performed using the official implementation \cite{brainscore} defined in the previous study \cite{schrimpf2018brain} with GPUs (NVIDIA V100, Tesla K80, and Tesla M60). Note that the neural response data and the images for recording them are included in the Brain-Score implementation (see the Brain-Score paper \cite{schrimpf2018brain} for a detailed description of these data).

\begin{figure}[!t]
  \centering
  \includegraphics[scale=1.0]{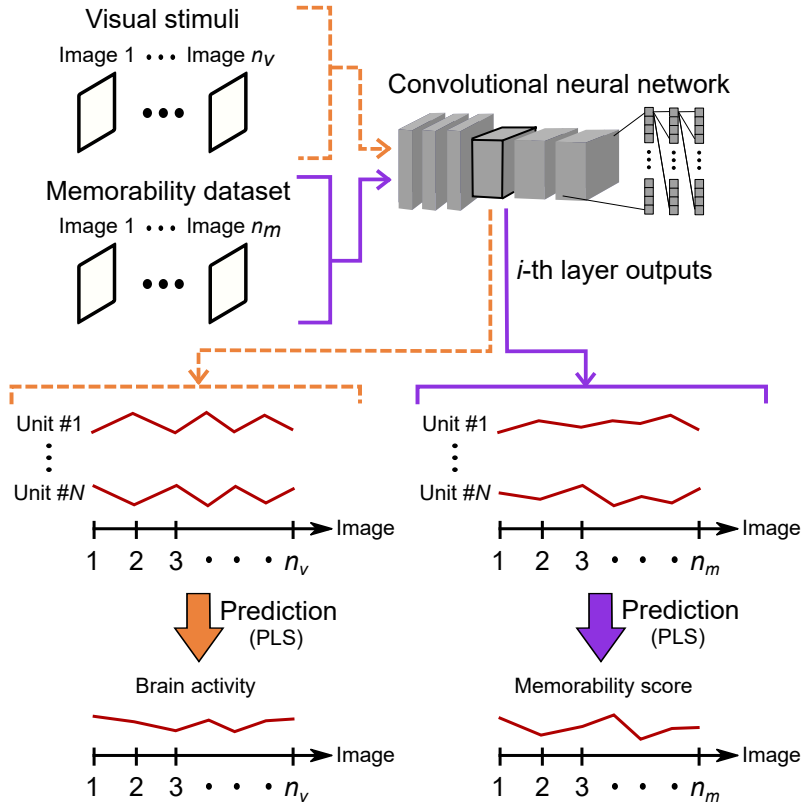}
  \caption{Evaluation of brain similarity and layer-wise memorability prediction accuracy in convolutional neural networks. Brain activity in each ventral stream (V1, V2, V4, or IT) region was predicted from each layer's outputs for the visual stimuli dataset using partial least squares (PLS) regression. The layer's brain similarity was defined as Pearson's correlation between brain activity and the predicted values. Similarly, using PLS regression, memorability scores for images were predicted from the layer outputs for the memorability dataset. The layer's memorability prediction accuracy was defined as Spearman's correlation between the memorability scores and predicted values.}
\end{figure}

\subsection{Layer-wise memorability prediction accuracy}
The image memorability scores were predicted with the outputs of each layer using PLS regression (depicted by the purple solid lines in Figure 1). In this regression, we set the number of latent variables to 25 and employed 5-fold cross-validation according to the official method for data splitting of the LaMem dataset \cite{ICCV15_Khosla}. Each layer's prediction accuracy was defined as Spearman's correlation coefficient between the image memorability scores and their predicted values using PLS regression. This analysis was performed using GPUs (NVIDIA A100, V100, and Tesla K80).

\subsection{Fine-tuning to memorability prediction}
Fine-tuning of the 64 CNN models pretrained for object recognition to image memorability prediction was also performed to further investigate memorability-related feature representations. In this regard, the last layer with an output size of 1,000 class labels in each CNN model was replaced by an output layer with a single unit. The LaMem image dataset \cite{ICCV15_Khosla} was used for training with 5-fold cross-validation. The training was performed for 100 epochs with the early stopping of 10 epochs using stochastic gradient descent with Nesterov momentum, a batch size of 32, an initial learning rate of $10^{-3}$, a momentum of 0.9, and GPUs (NVIDIA A100). The learning rate was divided by ten every ten epochs. We observed that the learning diverged in some models due to the high learning rates. In such cases, the learning rate was reset to one-tenth until the learning converged. We used a data augmentation scheme for ImageNet (resizing, cropping, and horizontal flipping) for the training dataset. Each model's performance for prediction accuracy was defined as Spearman's correlation coefficient between the image memorability scores and their predicted values. 

%
%
\section{Results}

\subsection{Brain similarity and layer-wise memorability prediction accuracy}
The correlation between brain similarity and memorability prediction accuracy in each layer across all ImageNet-pretrained CNN models was first investigated based on these findings. As a result, the memorability prediction accuracy showed the highest correlation with brain similarity with the IT cortex (V1, $\rho$ = 0.37; V2, $\rho$ = 0.76; V4, $\rho$ = 0.21; IT, $\rho$ = 0.82; Figure 2). This result shows that the layer containing memorability information also contains neuronal information in the IT cortex, suggesting that memorability-related feature representations are closely associated with those of the IT cortex.

\begin{figure}[!t]
  \centering
  \includegraphics[scale=0.85]{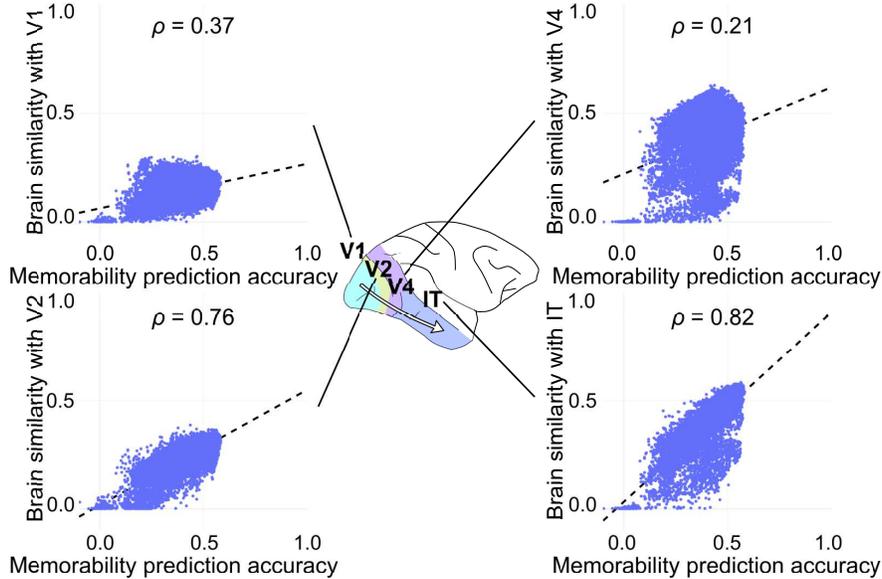}
  \caption{Correlation between brain similarity and layer-wise memorability prediction accuracy in ImageNet-pretrained CNN models. Vertical and horizontal axes show the brain similarity of a layer and the memorability prediction accuracy of a layer, respectively. Each point represents a layer, and there are 16,860 layers in total. There were low correlations in V1 and V4 (V1, Spearman's $\rho$ = 0.37; V4, $\rho$ = 0.21). The memorability prediction accuracy was moderately correlated with brain similarity with the V2 cortex ($\rho$ = 0.76). The highest correlation was observed in the inferior temporal (IT) cortex ($\rho$ = 0.82). The black dashed lines show the regression lines.}
\end{figure}

\subsection{Fine-tuning to memorability prediction}
The above results show that high memorability prediction accuracy is related to the feature representations of the IT cortex in models trained on object recognition data. This finding enables us to infer that CNN models fine-tuned to memorability prediction tend to acquire a feature representation similar to that of the IT cortex at the penultimate layer. Note that the penultimate layer outputs are the features directly used to predict memorability. To confirm the relationship, we fine-tuned the CNN models for object recognition to memorability prediction and examined the correlation between the prediction accuracy and the brain similarity of the penultimate layer with the IT cortex. The result showed that the memorability prediction accuracy of the fine-tuned CNN model correlated with the brain similarity with the IT cortex at the penultimate layer ($\rho$ = 0.56, $p < 0.001$, Figure 3). This result indicates that the model performance for memorability prediction relies on the feature representation acquisition of the IT cortex.

\begin{figure}[!t]
  \centering
  \includegraphics[scale=1.0]{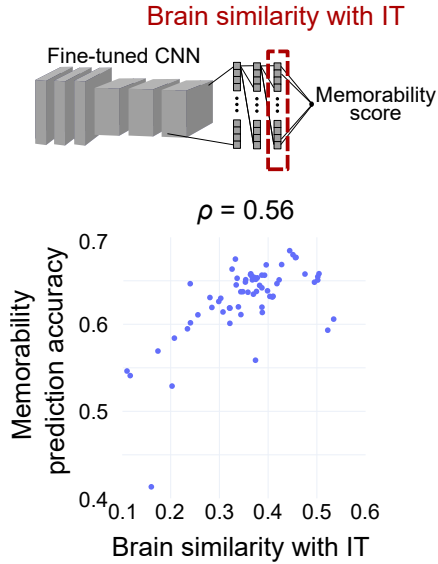}
  \caption{Memorability prediction accuracy of fine-tuned CNNs and brain similarity with the IT cortex at the penultimate layer of fine-tuned CNNs. Each point represents a CNN model, and there are 64 CNN models in total. The memorability prediction accuracy was evaluated using CNN models when all layers were fine-tuned on the LaMem dataset. CNN, convolutional neural network.}
\end{figure}

However, one might suspect this significant association in fine-tuned models was a pseudocorrelation mediated by object recognition performance before fine-tuning. In fact, model performance was positively correlated between object recognition and memorability prediction, as implied in a previous study \cite{Perera_2019_CVPR_Workshops} ($\rho$ = 0.46, $p < 0.001$, Figure S1), implying partial commonality between feature representations for object recognition and those for memorability prediction. Thus, the correlation between brain similarity with the IT cortex and object recognition accuracy in the ImageNet-pretrained CNN models was examined. The results showed that, unlike memorability prediction accuracy, object recognition accuracy was not positively correlated with brain similarity with the IT cortex at the penultimate layer ($\rho = -0.39$, ${\it p} < 0.001$, Figure 4A). On the contrary, a positive correlation between the best brain similarity with the IT cortex within the pretrained models and object recognition accuracy was found ($\rho$ = 0.57, $p < 0.001$, Figure 4B), which was consistent with a previous report \cite{schrimpf2018brain}. These results suggest that object recognition works by complex combinations in subsequent layers for IT representations acquired in the middle layer. In sum, memorability prediction is directly linked to the feature representations in the IT cortex.

Although not the main point of this paper, we found that DualPathNet92 \cite{chen2017dual} shows the highest prediction accuracy across the fine-tuned models and exceeds the accuracy of previous CNN models developed for memorability prediction (DualPathNet92, $\rho$ = 0.685; ResMem-Net \cite{praveen2021resmem}, $\rho$ = 0.679; AMNet \cite{fajtl2018amnet}, $\rho$ = 0.677; Table S2).

\begin{figure}[!t]
  \centering
  \includegraphics[scale=0.95]{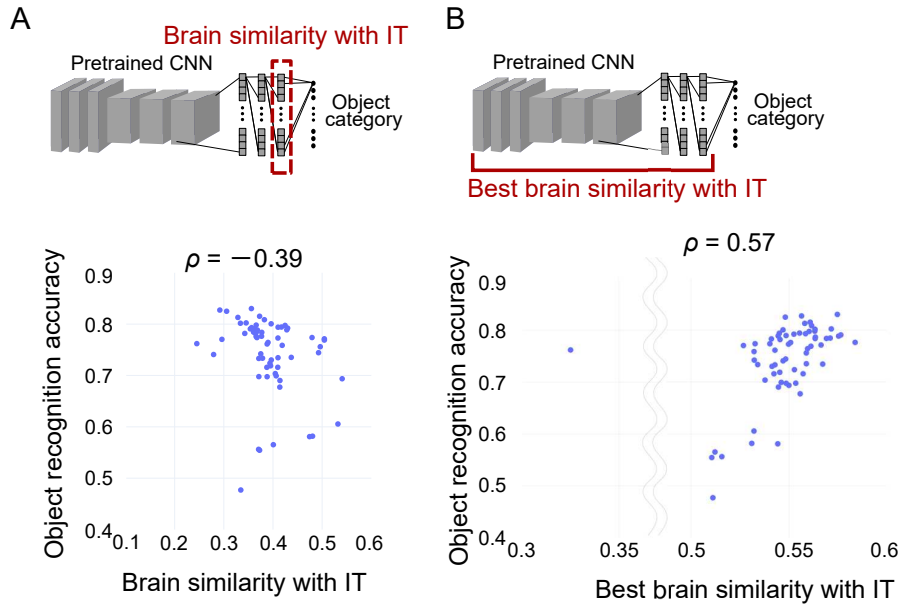}
  \caption{CNN performance for object recognition and brain similarity with the IT cortex. A, Object recognition accuracy of an ImageNet-pretrained CNN model and brain similarity with the IT cortex at the penultimate layer of the model. B, Object recognition accuracy of an ImageNet-pretrained CNN model and best brain similarity with the IT cortex among all layers in the model.}
\end{figure}

%
%
\section{Discussion}
Despite recent progress in predicting memorability using CNNs, it is still unclear which feature representations embedded in CNNs are crucial for high prediction performance. Here, from a biological perspective, feature representations of the IT cortex were demonstrated to be the key representations for predicting memorability. This finding suggests that recent success in predicting memorability using CNN models depends on acquiring visual representations in the IT cortex.

Our layer-wise analysis showed that IT cortex feature representations are critical for memorability prediction. This result is aligned with neuroscientific evidence demonstrating that neural population responses in the IT cortex are associated with memorability \cite{jaegle2019population}. In addition, our analysis demonstrated that feature representations in the V2 cortex are moderately correlated with memorability prediction accuracy. This result suggests that low-level image features acquired in the V2 cortex may also play an auxiliary role in memorability, although low-level image features are considered not to be associated with memorability \cite{6629991}. This finding may be consistent with a previous report suggesting that the V2 cortex is involved in object memory \cite{lopez2009role} Furthermore, analysis using fine-tuned models revealed that the memorability prediction accuracy is directly related to feature representations in the IT cortex. This result suggests that CNN models spontaneously acquire feature representations similar to those of the IT cortex through memorability learning and supports the relationship between feature representations for memorability and the IT cortex. Our data generalize or even extend the findings of previous research conducted by Jaegle et al. \cite{jaegle2019population}. Using a few CNN models trained for object recognition, they showed that as the layers become deep, the correlation between the L2 norm of the layer output and image memorability increases in strength. This finding seems consistent with the current finding that a feature representation similar to the IT cortex is critical to predicting image memorability with CNN models because deep (middle to back half) layers in CNN models trained for object recognition tend to acquire feature representations similar to the IT cortex \cite{cichy2016comparison, 10.1371/journal.pcbi.1003915, storrs2021diverse}. Our data more generally and comprehensively demonstrated the relationship between the CNN models' feature representation and memorability prediction accuracy.

Unlike the memorability prediction accuracy, the object recognition accuracy of the ImageNet-pretrained CNN models did not correlate with brain similarity with the IT cortex at the penultimate layer. However, the object recognition accuracy correlated with the best brain similarity with the IT cortex among the layers. Considering neuroscience evidence that IT cortex activity patterns are involved in object recognition \cite{dicarlo2012does}, this result may suggest that the object recognition function utilizing visual IT cortex representations needs additional subsequent processing in the later stages of visual processing. Thus, our results suggest that model performance for memorability and object recognition depends on IT cortex feature representations, but the strength of the relationship with the IT cortex differs for memorability and object recognition.

Although several studies focusing on prediction accuracy have proposed CNN models for memorability prediction, a model architecture that can more effectively acquire feature representations for memorability prediction has not been identified. In this regard, a previous study implied that a high-performance model for object recognition is a promising candidate for memorability prediction \cite{Perera_2019_CVPR_Workshops}. The relationship was investigated, and the correlation between object recognition accuracy and memorability prediction accuracy was confirmed ($\rho$ = 0.46 in Figure S1). From the perspective of brain similarity, a CNN model with a high brain similarity with the IT cortex at the penultimate layer was shown to tend to perform well in memorability prediction, and a correlation comparable to that between object recognition accuracy and memorability prediction accuracy was confirmed ($\rho$ = 0.56 in Figure 3). The overall results of this study suggest that building neural networks inspired by the way the brain works is just as important as building neural networks in an ad hoc fashion.

The limitations of this study should be addressed in future research. Brain similarity was evaluated using the electrode data of neuron activation. The electrode data precisely reflect local neuron activation. However, it is difficult to collect data from numerous locations (e.g., the data of the IT cortex in Brain-Score was collected from 168 sites \cite{schrimpf2018brain}). These limited data may not represent global patterns in the IT cortex. Thus, our results may be consistent only with local neuronal activity data in the IT cortex and diverge from global activity patterns in the IT cortex. To clarify this point, further analysis using electrode data for local neuronal activity and functional magnetic resonance imaging data for global patterns of neuronal activity would be helpful in future studies.

\section{Conclusions}
This study concluded that memorability relies on feature representations in the IT cortex. This study's findings suggest that model performance for memorability prediction depends on how well the model acquires feature representations embedded in the deeper region within the ventral visual stream, providing insight into the design of neuro-inspired models more comparable to human performance and the neural processing mechanism underlying visual cognitive functions in humans and artificial neural networks.

\section{Funding}
This research did not receive any specific grant from funding agencies in the public, commercial, or not-for-profit sectors.

\bibliography{mybibfile}

\begin{thebibliography}{10}
\expandafter\ifx\csname url\endcsname\relax
  \def\url#1{\texttt{#1}}\fi
\expandafter\ifx\csname urlprefix\endcsname\relax\def\urlprefix{URL }\fi
\expandafter\ifx\csname href\endcsname\relax
  \def\href#1#2{#2} \def\path#1{#1}\fi

\bibitem{ICCV15_Khosla}
A.~Khosla, A.~S. Raju, A.~Torralba, A.~Oliva, Understanding and predicting
  image memorability at a large scale, in: International conference on computer
  vision (ICCV), 2015.

\bibitem{6629991}
P.~Isola, J.~Xiao, D.~Parikh, A.~Torralba, A.~Oliva, What makes a photograph
  memorable?, IEEE transactions on pattern analysis and machine intelligence
  36~(7) (2014) 1469--1482.
\newblock \href {http://dx.doi.org/10.1109/TPAMI.2013.200}
  {\path{doi:10.1109/TPAMI.2013.200}}.

\bibitem{oliva2013makes}
A.~Oliva, P.~Isola, A.~Khosla, W.~A. Bainbridge, What makes a picture
  memorable?, SPIE Newsroom Article, 7 May 2013.

\bibitem{Perera_2019_CVPR_Workshops}
S.~Perera, A.~Tal, L.~Zelnik-Manor, Is image memorability prediction solved?,
  in: Proceedings of the IEEE/CVF conference on computer vision and pattern
  recognition (CVPR) workshops, 2019.

\bibitem{fajtl2018amnet}
J.~Fajtl, V.~Argyriou, D.~Monekosso, P.~Remagnino, Amnet: Memorability
  estimation with attention, in: Proceedings of the IEEE conference on computer
  vision and pattern recognition, 2018, pp. 6363--6372.

\bibitem{praveen2021resmem}
A.~Praveen, A.~Noorwali, D.~Samiayya, M.~Z. Khan, D.~R.~Vincent P M, A.~K. Bashir,
  V.~Alagupandi, Resmem-net: memory based deep cnn for image memorability
  estimation, PeerJ computer science 7 (2021) e767.

\bibitem{7410487}
R.~Dubey, J.~Peterson, A.~Khosla, M.-H. Yang, B.~Ghanem, What makes an object
  memorable?, in: 2015 IEEE international conference on computer vision (ICCV),
  2015, pp. 1089--1097.
\newblock \href {http://dx.doi.org/10.1109/ICCV.2015.130}
  {\path{doi:10.1109/ICCV.2015.130}}.

\bibitem{NEURIPS2019_70117ee3}
Z.~Li, W.~Brendel, E.~Walker, E.~Cobos, T.~Muhammad, J.~Reimer, M.~Bethge,
  F.~Sinz, Z.~Pitkow, A.~Tolias,
  \href{https://proceedings.neurips.cc/paper/2019/file/70117ee3c0b15a2950f1e82a215e812b-Paper.pdf}{Learning
  from brains how to regularize machines}, in: H.~Wallach, H.~Larochelle,
  A.~Beygelzimer, F.~d\textquotesingle Alch\'{e}-Buc, E.~Fox, R.~Garnett
  (Eds.), Advances in Neural Information Processing Systems, Vol.~32, Curran
  Associates, Inc., 2019.
\newline\urlprefix\url{https://proceedings.neurips.cc/paper/2019/file/70117ee3c0b15a2950f1e82a215e812b-Paper.pdf}

\bibitem{NEURIPS2020_98b17f06}
J.~Dapello, T.~Marques, M.~Schrimpf, F.~Geiger, D.~Cox, J.~J. DiCarlo,
  Simulating a primary visual cortex at the front of cnns improves robustness
  to image perturbations, in: Advances in Neural Information Processing
  Systems, Vol.~33, 2020, pp. 13073--13087.

\bibitem{yamins2014performance}
D.~L. Yamins, H.~Hong, C.~F. Cadieu, E.~A. Solomon, D.~Seibert, J.~J. DiCarlo,
  Performance-optimized hierarchical models predict neural responses in higher
  visual cortex, Proceedings of the National Academy of Sciences 111~(23)
  (2014) 8619--8624.

\bibitem{schrimpf2018brain}
M.~Schrimpf, J.~Kubilius, H.~Hong, N.~J. Majaj, R.~Rajalingham, E.~B. Issa,
  K.~Kar, P.~Bashivan, J.~Prescott-Roy, K.~Schmidt, et~al., Brain-score: Which
  artificial neural network for object recognition is most brain-like?, bioRxiv
  (2018) 407007.

\bibitem{dicarlo2012does}
J.~J. DiCarlo, D.~Zoccolan, N.~C. Rust, How does the brain solve visual object
  recognition?, Neuron 73~(3) (2012) 415--434.

\bibitem{jaegle2019population}
A.~Jaegle, V.~Mehrpour, Y.~Mohsenzadeh, T.~Meyer, A.~Oliva, N.~Rust, Population
  response magnitude variation in inferotemporal cortex predicts image
  memorability, eLife 8 (2019) e47596.

\bibitem{deng2009imagenet}
J.~Deng, W.~Dong, R.~Socher, L.-J. Li, K.~Li, L.~Fei-Fei, Imagenet: A
  large-scale hierarchical image database, in: IEEE Conference on Computer
  Vision and Pattern Recognition, 2009, pp. 248--255.

\bibitem{krizhevsky2012imagenet}
A.~Krizhevsky, I.~Sutskever, G.~E. Hinton, Imagenet classification with deep
  convolutional neural networks, in: Advances in Neural Information Processing
  Systems, 2012, pp. 1097--1105.

\bibitem{simonyan2014very}
K.~Simonyan, A.~Zisserman, Very deep convolutional networks for large-scale
  image recognition, arXiv (2015) 1409.1556.

\bibitem{he2016deep}
K.~He, X.~Zhang, S.~Ren, J.~Sun, Deep residual learning for image recognition,
  in: Conference on Computer Vision and Pattern Recognition, 2016, pp.
  770--778.

\bibitem{huang2017densely}
G.~Huang, Z.~Liu, K.~Q. Weinberger, L.~van~der Maaten, Densely connected
  convolutional networks, in: Conference on computer vision and pattern
  recognition, Vol.~1, 2017, pp. 4700--4708.

\bibitem{Hu_2018_CVPR}
J.~Hu, L.~Shen, G.~Sun, Squeeze-and-excitation networks, in: Conference on
  Computer Vision and Pattern Recognition, 2018.

\bibitem{kubilius2018cornet}
J.~Kubilius, M.~Schrimpf, A.~Nayebi, D.~Bear, D.~L. Yamins, J.~J. DiCarlo,
  Cornet: modeling the neural mechanisms of core object recognition, bioRxiv
  (2018) 408385.

\bibitem{pytorch}
Pytorch library,
  \url{https://github.com/pytorch/vision/tree/master/torchvision/models}
  (accessed 20 February 2020).

\bibitem{cadena}
R.~Cadena, Pretrained models for {P}ytorch,
  \url{https://github.com/Cadene/pretrained-models.pytorch} (accessed 20
  February 2020).

\bibitem{cornet}
M.~Schrimpf, {COR}net: Modeling the neural mechanisms of core object
  recognition, \url{https://github.com/dicarlolab/CORnet} (accessed 17 February
  2020).

\bibitem{eff}
L.~Melas-Kyriazi, {EfficientNet PyTorch},
  \url{https://github.com/lukemelas/EfficientNet-PyTorch} (accessed 14 May
  2020).

\bibitem{goetschalckx2019memcat}
L.~Goetschalckx, J.~Wagemans, Memcat: a new category-based image set quantified
  on memorability, PeerJ 7 (2019) e8169.

\bibitem{brainscore}
M.~Schrimpf, {Brain-Score}, \url{https://github.com/brain-score/brain-score}
  (accessed 13 May 2020).

\bibitem{chen2017dual}
Y.~Chen, J.~Li, H.~Xiao, X.~Jin, S.~Yan, J.~Feng, Dual path networks, in:
  Advances in Neural Information Processing Systems, 2017, pp. 4467--4475.

\bibitem{lopez2009role}
M.~F. L{\'o}pez-Aranda, J.~F. L{\'o}pez-T{\'e}llez, I.~Navarro-Lobato,
  M.~Masmudi-Mart{\'\i}n, A.~Guti{\'e}rrez, Z.~U. Khan, Role of layer 6 of {V2}
  visual cortex in object-recognition memory, Science 325~(5936) (2009) 87--89.

\bibitem{cichy2016comparison}
R.~M. Cichy, A.~Khosla, D.~Pantazis, A.~Torralba, A.~Oliva, Comparison of deep
  neural networks to spatio-temporal cortical dynamics of human visual object
  recognition reveals hierarchical correspondence, Scientific Reports 6~(1)
  (2016) 1--13.

\bibitem{10.1371/journal.pcbi.1003915}
S.-M. Khaligh-Razavi, N.~Kriegeskorte,
  \href{https://doi.org/10.1371/journal.pcbi.1003915}{Deep supervised, but not
  unsupervised, models may explain it cortical representation}, PLOS
  Computational Biology 10~(11) (2014) 1--29.
\newblock \href {http://dx.doi.org/10.1371/journal.pcbi.1003915}
  {\path{doi:10.1371/journal.pcbi.1003915}}.
\newline\urlprefix\url{https://doi.org/10.1371/journal.pcbi.1003915}

\bibitem{storrs2021diverse}
K.~R. Storrs, T.~C. Kietzmann, A.~Walther, J.~Mehrer, N.~Kriegeskorte, Diverse
  deep neural networks all predict human inferior temporal cortex well, after
  training and fitting, Journal of cognitive neuroscience 33~(10) (2021)
  2044--2064.

\end{thebibliography}


\begin{thebibliography}{10}
\expandafter\ifx\csname url\endcsname\relax
  \def\url#1{\texttt{#1}}\fi
\expandafter\ifx\csname urlprefix\endcsname\relax\def\urlprefix{URL }\fi
\expandafter\ifx\csname href\endcsname\relax
  \def\href#1#2{#2} \def\path#1{#1}\fi

\bibitem{krizhevsky2012imagenet}
A.~Krizhevsky, I.~Sutskever, G.~E. Hinton, Imagenet classification with deep
  convolutional neural networks, in: Advances in Neural Information Processing
  Systems, 2012, pp. 1097--1105.

\bibitem{ioffe2015batch}
S.~Ioffe, C.~Szegedy, Batch normalization: accelerating deep network training
  by reducing internal covariate shift, arXiv (2015) 1502.03167.

\bibitem{jia2014caffe}
Y.~Jia, E.~Shelhamer, J.~Donahue, S.~Karayev, J.~Long, R.~Girshick,
  S.~Guadarrama, T.~Darrell, Caffe: convolutional architecture for fast feature
  embedding, arXiv (2014) 1408.5093.

\bibitem{kubilius2018cornet}
J.~Kubilius, M.~Schrimpf, A.~Nayebi, D.~Bear, D.~L. Yamins, J.~J. DiCarlo,
  Cornet: modeling the neural mechanisms of core object recognition, bioRxiv
  (2018) 408385.

\bibitem{kubilius2019brain}
J.~Kubilius, M.~Schrimpf, K.~Kar, R.~Rajalingham, et~al., Brain-like object
  recognition with high-performing shallow recurrent anns, in: Advances in
  Neural Information Processing Systems, 2019, pp. 12785--12796.

\bibitem{huang2017densely}
G.~Huang, Z.~Liu, K.~Q. Weinberger, L.~van~der Maaten, Densely connected
  convolutional networks, in: Conference on computer vision and pattern
  recognition, Vol.~1, 2017, pp. 4700--4708.

\bibitem{chen2017dual}
Y.~Chen, J.~Li, H.~Xiao, X.~Jin, S.~Yan, J.~Feng, Dual path networks, in:
  Advances in Neural Information Processing Systems, 2017, pp. 4467--4475.

\bibitem{tan2019efficientnet}
M.~Tan, Q.~V. Le, Efficientnet: Rethinking model scaling for convolutional
  neural networks, arXiv (2019) 1905.11946.

\bibitem{git0000}
fb.resnet.torch, \url{https://github.com/facebookarchive/fb.resnet.torch}
  (accessed 20 February 2020).

\bibitem{Szegedy_2015_CVPR}
C.~Szegedy, W.~Liu, Y.~Jia, P.~Sermanet, S.~Reed, D.~Anguelov, D.~Erhan,
  V.~Vanhoucke, A.~Rabinovich, Going deeper with convolutions, in: Conference
  on Computer Vision and Pattern Recognition, 2015.

\bibitem{szegedy2016inc}
C.~Szegedy, S.~Ioffe, V.~Vanhoucke, A.~A. Alemi, Inception-v4, inception-resnet
  and the impact of residual connections on learning, in: International
  Conference Learning Representations Workshop, 2016.

\bibitem{szegedy2016rethinking}
C.~Szegedy, V.~Vanhoucke, S.~Ioffe, J.~Shlens, Z.~Wojna, Rethinking the
  inception architecture for computer vision, in: Conference on computer vision
  and pattern recognition, 2016, pp. 2818--2826.

\bibitem{sandler2018mobilenetv2}
M.~Sandler, A.~Howard, M.~Zhu, A.~Zhmoginov, L.-C. Chen, Mobilenetv2: inverted
  residuals and linear bottlenecks, in: Conference on computer vision and
  pattern recognition, 2018, pp. 4510--4520.

\bibitem{Zoph_2018_CVPR}
B.~Zoph, V.~Vasudevan, J.~Shlens, Q.~V. Le, Learning transferable architectures
  for scalable image recognition, in: Conference on Computer Vision and Pattern
  Recognition, 2018.

\bibitem{tan2019mnasnet}
M.~Tan, B.~Chen, R.~Pang, V.~Vasudevan, M.~Sandler, A.~Howard, Q.~V. Le,
  Mnasnet: platform-aware neural architecture search for mobile, in:
  Proceedings of the IEEE Conference on Computer Vision and Pattern
  Recognition, 2019, pp. 2820--2828.

\bibitem{liu2018progressive}
C.~Liu, B.~Zoph, M.~Neumann, J.~Shlens, W.~Hua, L.-J. Li, L.~Fei-Fei,
  A.~Yuille, J.~Huang, K.~Murphy, Progressive neural architecture search, in:
  Proceedings of the European Conference on Computer Vision, 2018, pp. 19--34.

\bibitem{zhang2017polynet}
X.~Zhang, Z.~Li, C.~Change~Loy, D.~Lin, Polynet: a pursuit of structural
  diversity in very deep networks, in: Proceedings of the IEEE Conference on
  Computer Vision and Pattern Recognition, 2017, pp. 718--726.

\bibitem{he2016deep}
K.~He, X.~Zhang, S.~Ren, J.~Sun, Deep residual learning for image recognition,
  in: Conference on Computer Vision and Pattern Recognition, 2016, pp.
  770--778.

\bibitem{xie2017aggregated}
S.~Xie, R.~Girshick, P.~Doll{\'a}r, Z.~Tu, K.~He, Aggregated residual
  transformations for deep neural networks, in: Conference on Computer Vision
  and Pattern Recognition, 2017, pp. 5987--5995.

\bibitem{Hu_2018_CVPR}
J.~Hu, L.~Shen, G.~Sun, Squeeze-and-excitation networks, in: Conference on
  Computer Vision and Pattern Recognition, 2018.

\bibitem{Zhang_2018_CVPR}
X.~Zhang, X.~Zhou, M.~Lin, J.~Sun, Shufflenet: an extremely efficient
  convolutional neural network for mobile devices, in: Conference on Computer
  Vision and Pattern Recognition, 2018.

\bibitem{iandola2016squeezenet}
F.~N. Iandola, S.~Han, M.~W. Moskewicz, K.~Ashraf, W.~J. Dally, K.~Keutzer,
  Squeezenet: Alexnet-level accuracy with 50x fewer parameters and
  \textless0.5{MB} model size, arXiv (2016) 1602.07360.

\bibitem{simonyan2014very}
K.~Simonyan, A.~Zisserman, Very deep convolutional networks for large-scale
  image recognition, arXiv (2015) 1409.1556.

\bibitem{zagoruyko2016wide}
S.~Zagoruyko, N.~Komodakis, Wide residual networks, arXiv (2016) 1605.07146.

\bibitem{Chollet_2017_CVPR}
F.~Chollet, Xception: deep learning with depthwise separable convolutions, in:
  Conference on Computer Vision and Pattern recognition, 2017, pp. 1251--1258.

\bibitem{ICCV15_Khosla}
A.~Khosla, A.~S. Raju, A.~Torralba, A.~Oliva, Understanding and predicting
  image memorability at a large scale, in: International conference on computer
  vision (ICCV), 2015.

\bibitem{basavaraju2019multiple}
S.~Basavaraju, A.~Sur, Multiple instance learning based deep cnn for image
  memorability prediction, Multimedia tools and applications 78~(24) (2019)
  35511--35535.

\bibitem{Perera_2019_CVPR_Workshops}
S.~Perera, A.~Tal, L.~Zelnik-Manor, Is image memorability prediction solved?,
  in: Proceedings of the IEEE/CVF conference on computer vision and pattern
  recognition (CVPR) workshops, 2019.

\bibitem{fajtl2018amnet}
J.~Fajtl, V.~Argyriou, D.~Monekosso, P.~Remagnino, Amnet: Memorability
  estimation with attention, in: Proceedings of the IEEE conference on computer
  vision and pattern recognition, 2018, pp. 6363--6372.

\bibitem{praveen2021resmem}
A.~Praveen, A.~Noorwali, D.~Samiayya, M.~Z. Khan, D.~R.~V. PM, A.~K. Bashir,
  V.~Alagupandi, Resmem-net: memory based deep cnn for image memorability
  estimation, PeerJ computer science 7 (2021) e767.

\end{thebibliography}

\end{document}


\begin{center}
{\Large {\bf Supplementary material}\\ Feature representations useful for predicting image memorability}
\end{center}

\begin{center}
Takumi Harada, Hiroyuki Sakai
\end{center}

\begin{table}[hb]
  \caption{CNN model list.}
  \label{model-table}
  \centering
  \begin{tabular}{ll}
    \toprule
    \multicolumn{2}{c}{{\bf model}}                   \\
    \midrule
    AlexNet\cite{krizhevsky2012imagenet}& BNInception\cite{ioffe2015batch}\\ CaffeResNet101\cite{jia2014caffe}& CORnet-R\cite{kubilius2018cornet}\\ CORnet-RT\cite{kubilius2018cornet}& CORnet-S\cite{kubilius2019brain}\\ CORnet-Z\cite{kubilius2018cornet}& DenseNet121\cite{huang2017densely}\\ DenseNet161\cite{huang2017densely}& DenseNet169\cite{huang2017densely}\\ DenseNet201\cite{huang2017densely}& DualPathNet68\cite{chen2017dual}\\ DualPathNet68b\cite{chen2017dual}& DualPathNet92\cite{chen2017dual}\\ DualPathNet98\cite{chen2017dual}& DualPathNet107\cite{chen2017dual}\\ DualPathNet113\cite{chen2017dual}& EfficientNet B0\cite{tan2019efficientnet}\\
    EfficientNet B1\cite{tan2019efficientnet}&
    EfficientNet B2\cite{tan2019efficientnet}\\
    EfficientNet B3\cite{tan2019efficientnet}&
    EfficientNet B4\cite{tan2019efficientnet}\\
    FBResNet152\cite{git0000}& GoogLeNet\cite{Szegedy_2015_CVPR}\\ InceptionResNetV2\cite{szegedy2016inc}& InceptionV3\cite{szegedy2016rethinking}\\ InceptionV4\cite{szegedy2016inc}& MobileNet v2\cite{sandler2018mobilenetv2}\\ NASNet-A-Large\cite{Zoph_2018_CVPR}& NASNet-A-Mobile\cite{Zoph_2018_CVPR}\\ MNASNet (with depth multiplier of 0.5)\cite{tan2019mnasnet}&
    MNASNet (with depth multiplier of 1.0)\cite{tan2019mnasnet}\\
    PNASNet-5-Large\cite{liu2018progressive}& PolyNet\cite{zhang2017polynet}\\ ResNet101\cite{he2016deep}& ResNet152\cite{he2016deep}\\ ResNet18\cite{he2016deep}& ResNet34\cite{he2016deep}\\ ResNet50\cite{he2016deep}& ResNeXt50 32x4d\cite{xie2017aggregated}\\ ResNeXt 101 32x4d\cite{xie2017aggregated}& ResNeXt-101 32x8d\cite{xie2017aggregated}\\ ResNeXt101 64x4d\cite{xie2017aggregated}& SENet154\cite{Hu_2018_CVPR}\\ SE-ResNet50\cite{Hu_2018_CVPR}& SE-ResNet101\cite{Hu_2018_CVPR}\\ SE-ResNet152\cite{Hu_2018_CVPR}& SE-ResNeXt50 32x4d\cite{Hu_2018_CVPR}\\ SE-ResNeXt101 32x4d\cite{Hu_2018_CVPR}& ShuffleNet v2 (with 0.5x output channels)\cite{Zhang_2018_CVPR}\\ShuffleNet v2 (with 1.0x output channels)\cite{Zhang_2018_CVPR}& SqueezeNet1\_0\cite{iandola2016squeezenet}\\ SqueezeNet1\_1\cite{iandola2016squeezenet}& VGG11\cite{simonyan2014very}\\ VGG13\cite{simonyan2014very}& VGG16\cite{simonyan2014very}\\ VGG19\cite{simonyan2014very}& VGG11\_BN\cite{simonyan2014very}\\ VGG13\_BN\cite{simonyan2014very}& VGG16\_BN\cite{simonyan2014very}\\ VGG19\_BN\cite{simonyan2014very}& Wide ResNet-50-2\cite{zagoruyko2016wide} \\ Wide ResNet-101-2\cite{zagoruyko2016wide}& Xception\cite{Chollet_2017_CVPR} \\
    \bottomrule
  \end{tabular}
\end{table}

\begin{figure}
  \centering
  \includegraphics[scale=1.0]{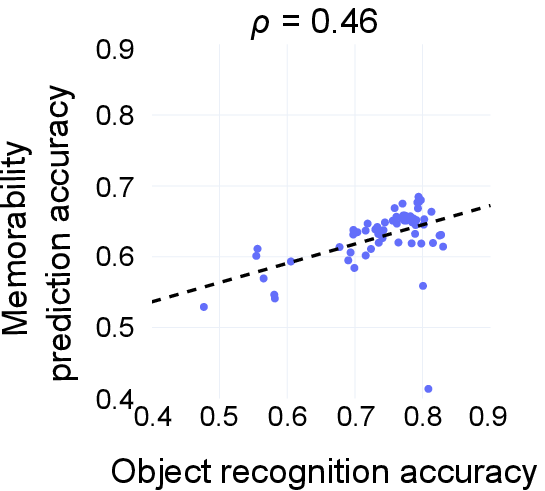}
  \caption{Each model performance for object recognition and memorability prediction. The object recognition accuracy was evaluated using CNN models trained on ImageNet dataset. The memorability prediction accuracy was evaluated using CNN models when all the layers are fine-tuned on LaMem dataset. The black dashed line shows the regression line. }
\end{figure}

\begin{table}[h]
\caption{Memorability prediction accuracy in previously proposed models.}
\label{table:data_type}
\centering
\begin{tabular}{cc}
\hline
model  & accuracy\\
\hline 
MemNet\cite{ICCV15_Khosla}  &0.64 \\
MCDRNet\cite{basavaraju2019multiple}  &0.663 \\
MemBoost\cite{Perera_2019_CVPR_Workshops} & 0.67\\
EMNet\cite{basavaraju2019multiple}  & 0.671 \\
AMNet\cite{fajtl2018amnet}  &    0.677 \\
ResMem-Net\cite{praveen2021resmem} & 0.679\\
DualPathNet92 & 0.685\\
\hline
\end{tabular}
\end{table}

\clearpage

\bibliography{mybibfile2}